\pdfoutput=1

\documentclass[11pt]{article}

\usepackage{naacl2021}

\usepackage{times}
\usepackage{latexsym}
\usepackage{colortbl}
\usepackage[T1]{fontenc}

\usepackage[utf8]{inputenc}

\usepackage{microtype}

\usepackage{algorithm}
\usepackage{algorithmic}
\usepackage{graphics}
\usepackage{graphicx}
\usepackage{booktabs}
\usepackage{multirow}
\usepackage{makecell}
\usepackage{caption}

\usepackage{boxedminipage}
\usepackage{wrapfig,lipsum,booktabs}
\usepackage{times}
\usepackage{longtable}
\usepackage{amsmath}
\usepackage{subcaption}
\usepackage{tabularx}
\usepackage{colortbl}
\usepackage{latexsym}
\usepackage{makecell}
\usepackage{graphicx}
\usepackage{caption}
\usepackage{multirow}
\usepackage{makecell}

\newcommand{\B}[1] {\boldsymbol{#1}}

\def\bW{{\B{W}}}

\def\be{{\B{e}}}

\def\bp{{\B{p}}}

\def\bs{{\B{s}}}

\def\bu{{\B{u}}}

\def\by{{\B{y}}}
\def\bz{{\B{z}}}

\makeatletter
\newcommand{\thickhline}{%
    \noalign {\ifnum 0=`}\fi \hrule height 1pt
    \futurelet \reserved@a \@xhline
}

\newcommand{\thickerhline}{%
    \noalign {\ifnum 0=`}\fi \hrule height 2pt
    \futurelet \reserved@a \@xhline
}

\makeatother

\title{Enhancing Factual Consistency of Abstractive Summarization}

\author{Chenguang Zhu$^1$, William Hinthorn$^1$, Ruochen Xu$^1$, Qingkai Zeng$^2$, \\
\textbf{Michael Zeng}$^1$, \textbf{Xuedong Huang}$^1$, \textbf{Meng Jiang}$^2$ \\
\textsuperscript{\rm 1} Microsoft Cognitive Services Group\\
\textsuperscript{\rm 2} University of Notre Dame \\
\texttt{\{chezhu,wihintho,ruox,nzeng,xdh\}@microsoft.com,\{qzeng,mjiang2\}@nd.edu}
}

\begin{document}
\maketitle

\begin{abstract}
Automatic abstractive summaries are found to often distort or fabricate facts in the article. This inconsistency between summary and original text has seriously impacted its applicability. 
We propose a fact-aware summarization model \textsc{FASum} to extract and integrate factual relations into the summary generation process via graph attention. We then design a factual corrector model FC to automatically correct factual errors from summaries generated by existing systems. Empirical results\footnote{We provide the prediction results of all models at \url{https://github.com/zcgzcgzcg1/FASum/}.} show that the fact-aware summarization can produce abstractive summaries with higher factual consistency compared with existing systems, and the correction model improves the factual consistency of given summaries via modifying only a few keywords.
\end{abstract}

\section{Introduction}
\label{sec:intro}

Text summarization models aim to produce an abridged version of long text while preserving salient information. Abstractive summarization is a type of such models that can freely generate summaries, with no constraint on the words or phrases used. This format is closer to human-edited summaries and is both flexible and informative. Thus, there are numerous approaches to produce abstractive summaries \citep{pgnet,drm,unilm,bottomup}.

\definecolor{darkpastelgreen}{rgb}{0.01, 0.75, 0.24}
\definecolor{seagreen}{rgb}{0.18, 0.55, 0.34}
\definecolor{mountainmeadow}{rgb}{0.19, 0.73, 0.56}
\definecolor{forestgreen}{rgb}{0.13, 0.55, 0.13}
\definecolor{fireenginered}{rgb}{0.81, 0.09, 0.13}

\begin{table}[t]
\centering
\small
\begin{tabular}{c|l}
\thickhline
Article & \makecell[l]{
\textbf{Real Madrid ace Gareth Bale} treated \\
himself to a Sunday evening BBQ... \\
The Welsh wizard was ... \textbf{scoring twice} and \\
assisting another in an impressive victory...\\
\textbf{Cristiano Ronaldo scored five goals} against \\
Granada on Sunday ...}\\ 
\hline 
\hline
\textsc{BottomUp} & \makecell[l]{... \textbf{\textcolor{fireenginered}{The Real Madrid ace}} scored five goals \\
against Granada on Sunday. The Welsh \\
wizard was in impressive form for...}\\
\hline
\textsc{Seq2Seq} & \makecell[l]{...\textbf{\textcolor{fireenginered}{Gareth Bale scored five}} and assisted \\another in an impressive win in Israel...} \\
\hline
\makecell[c]{\textsc{FASum}\\(Ours)} & \makecell[l]{...\textbf{\textcolor{forestgreen}{Gareth Bale scored twice}} and helped his \\
side to a sensational 9-1 win. \textbf{\textcolor{forestgreen}{Cristiano}} \\
\textbf{\textcolor{forestgreen}{Ronaldo scored five goals}} against Granada \\
on Sunday...} \\
\thickhline
\end{tabular}
\caption{Example article and summary excerpts from CNN/DailyMail dataset.} \label{table:first_example}
\end{table}

However, one prominent issue with abstractive summarization is factual inconsistency. It refers to the \textit{hallucination} phenomenon that the summary sometimes distorts or fabricates the facts in the article. 
Recent studies show that up to 30\% of the summaries generated by abstractive models contain such factual inconsistencies \citep{eval,ranking}, raising concerns about the credibility and usability of these systems. Table~\ref{table:first_example} demonstrates an example article and excerpts of generated summaries. As shown, the article mentions that Real Madrid ace Gareth Bale scored twice and Cristiano Ronaldo scored five goals. However, both \textsc{BottomUp} \citep{bottomup} and \textsc{\textsc{Seq2Seq}} wrongly states that Bale scored five goals. Comparatively, our model \textsc{FASum} generates a summary that correctly exhibits the fact in the article. And as shown in Section~\ref{sec:novel}, our model achieves higher factual consistency not just by making more copies from the article.

\begin{table*}[t]
\centering
\small
\begin{tabular}{c|l}
\thickhline
Article & \makecell[l]{The flame of remembrance burns in Jerusalem, and a song of memory haunts Valerie Braham as it never has \\ before. This year, \textbf{Israel's Memorial Day} commemoration is for bereaved family members such as Braham. \\``Now I truly understand everyone who has lost a loved one," Braham said. Her husband, \textbf{Philippe Braham}, \\ was one of 17 people killed in January's terror attacks in Paris...\\
As Israel mourns on the nation's remembrance day, \textbf{French Prime Minister} Manuel Valls announced after \\ his weekly Cabinet meeting that French authorities had foiled a terror plot...
}\\
\hline
\hline
\textsc{BottomUp} & \makecell[l]{\textbf{\textcolor{fireenginered}{Valerie Braham}} was one of 17 people killed in January 's terror attacks in Paris. \textbf{\textcolor{fireenginered}{France's memorial day}} \\ commemoration is for bereaved family members as Braham. \textbf{\textcolor{fireenginered}{Israel's Prime Minister}} says the terror plot \\ has not been done.}\\
\hline
\makecell[c]{Corrected by\\\textsc{FC}} & \makecell[l]{\textbf{\textcolor{forestgreen}{Philippe Braham}} was one of 17 people killed in January's terror attacks in Paris. \textbf{\textcolor{forestgreen}{Israel's memorial day}}\\commemoration is for bereaved family members as Braham. \textbf{\textcolor{forestgreen}{France's Prime Minister}} says the terror plot\\ has not been done.} \\
\thickhline
\end{tabular}
\caption{Example excerpts of an article from CNN/DailyMail and the summary generated by \textsc{BottomUp}. Factual errors are marked in red. The correction made by our model \textsc{FC} are marked in green.} 
\label{table:fc_example}
\end{table*}


On the other hand, most existing abstractive summarization models apply a conditional language model to focus on the token-level accuracy of summaries, while neglecting semantic-level consistency between the summary and article. Therefore, the generated summaries are often high in token-level metrics like ROUGE \citep{rouge} but lack factual consistency. In view of this, we argue that a robust abstractive summarization system must be equipped with factual knowledge to accurately summarize the article. 

In this paper, we represent facts in the form of knowledge graphs. Although there are numerous efforts in building commonly applicable knowledge graphs such as ConceptNet \citep{conceptnet}, we find that these tools are more useful in conferring commonsense knowledge. In abstractive summarization for contents like news articles, many entities and relations are previously unseen. Plus, our goal is to produce summaries that do not conflict with the facts in the article. Thus, we propose to extract factual knowledge from the article itself.

We employ the information extraction (IE) tool OpenIE \citep{openie} to extract facts from the article in the form of relational tuples: (subject, relation, object). 
This graph contains the facts in the article and is integrated in the summary generation process.

Then, we use a graph attention network \citep{gat} to obtain the representation of each node, and fuse that into a transformer-based encoder-decoder architecture via attention. We denote this model as the Fact-Aware Summarization model, \textsc{FASum}.



In addition, to be generally applicable for all existing summarization systems, we propose a Factual Corrector model, \textsc{FC}, to help improve the factual consistency of any given summary. We frame the correction process as a seq2seq problem: the input is the original summary and the article, and the output is the corrected summary. \textsc{FC} has the same architecture as UniLM \citep{unilm} and initialized with weights from RoBERTa-Large \citep{roberta}. We finetune it as a denoising autoencoder. The training data is synthetically generated via randomly replacing entities in the ground-truth summary with wrong ones in the article. As shown in Table~\ref{table:fc_example}, \textsc{FC} makes three corrections, replacing the original wrong entities which appear elsewhere in the article with the right ones.

In the experiments, we leverage an independently trained BERT-based \citep{devlin2018bert} factual consistency evaluator \citep{eval}. Results show that on CNN/DailyMail, \textsc{FASum} obtains 0.6\% higher fact consistency scores than \textsc{UniLM} \citep{unilm} and 3.9\% higher than \textsc{BottomUp} \citep{bottomup}. Moreover, after correction by \textsc{FC}, the factual score of summaries from \textsc{BottomUp} increases 1.4\% on CNN/DailyMail and 0.9\% on XSum, and the score of summaries from \textsc{TConvS2S} increases 3.1\% on XSum. We also conduct human evaluation to verify the effectiveness of our models.

We further propose an easy-to-compute model-free metric, relation matching rate (RMR), to evaluate factual consistency given a summary and the article. This metric employs the extracted relations and does not require human-labelled summaries. 
Under this metric, we show that our models can help enhance the factual consistency of summaries.








\section{Related Work}
\label{sec:related}
\subsection{Abstractive Summarization}
Abstractive text summarization has been intensively studied in recent literature. 
\citet{nnlm} introduces an attention-based seq2seq model for abstractive sentence summarization. \citet{pgnet} uses copy-generate mechanism that can both produce words from the vocabulary via a generator and copy words from the article via a pointer. \citet{drm} leverages reinforcement learning to improve summarization quality. \citet{bottomup} uses a content selector to over-determine phrases in source documents that helps constrain the model to likely phrases. \citet{leadbias} defines a pretraining scheme for summarization and produces a zero-shot abstractive summarization model. \citet{unilm} employs different masking techniques for both NLU and NLG tasks, resulting in the \textsc{UniLM} model. 
\citet{bart} employs denoising techniques to help generation tasks including summarization.

\subsection{Fact-Aware Summarization}
Entailment models have been used to evaluate and enhance factual consistency of summarization. \citet{ensure} co-trains summarization and entailment and employs an entailment-aware decoder. \citet{ranking} proposes using off-the-shelf entailment models to rerank candidate summary sentences to boost factual consistency. 

\citet{yuhao} employs descriptor vectors to improve factual consistency in medical summarization. 
\citet{furu} extracts relational information from the article and maps it to a sequence as an additional input to the encoder. \citet{beliz} employs an entity-aware transformer structure for knowledge integration, and \citet{headline} improves factual consistency of generated headlines by filtering out training data with more factual errors. 
In comparison, our model utilizes the knowledge graph extracted from the article and fuses it into the generated text via neural graph computation.

To correct factual errors, \citet{multifact} uses pre-trained NLU models to rectify one or more wrong entities in the summary. Concurrent to our work, \citet{factcorrect} employs the generation model BART \citep{bart} to produce corrected summaries.

Several approaches have been proposed to evaluate a summary’s factual consistency \citep{kryscinski-etal-2019-neural,assessfactual,factualfaithful}. \citet{bertscore} employs BERT to compute similarity between pairs of words in the summary and article. \citet{qametric,feqa} use question answering accuracy to measure factual consistency. 
\citet{eval} applies various transformations on the summary to produce training data for a BERT-based classification model, FactCC, which shows a high correlation with human metrics. Therefore, we use FactCC as the factual evaluator in this paper.


\section{Model}
\label{sec:model}
\subsection{Problem Formulation}
We formalize abstractive summarization as a supervised seq2seq problem. The input consists of $a$ pairs of articles and summaries: $\{(X_1, Y_1), (X_2, Y_2), ..., (X_a, Y_a)\}$. Each article is tokenized into $X_i=(x_1,...,x_{L_i})$ and each summary is tokenized into $Y_i=(y_1,...,y_{N_i})$. In abstrative summarization, the model-generated summary can contain tokens, phrases and sentences not present in the article. For simplicity, in the following we will drop the data index subscript. Therefore, each training pair becomes $X=(x_1,...,x_m), Y=(y_1,...,y_n)$, and the model needs to generate an abstrative summary $\hat{Y}=(\hat{y}_1,...,\hat{y}_{n'})$.

\begin{figure*}[thbp]
\centering
\vspace{2.8\baselineskip}
\includegraphics[scale=0.5,trim=6cm 6cm 6cm 4cm]{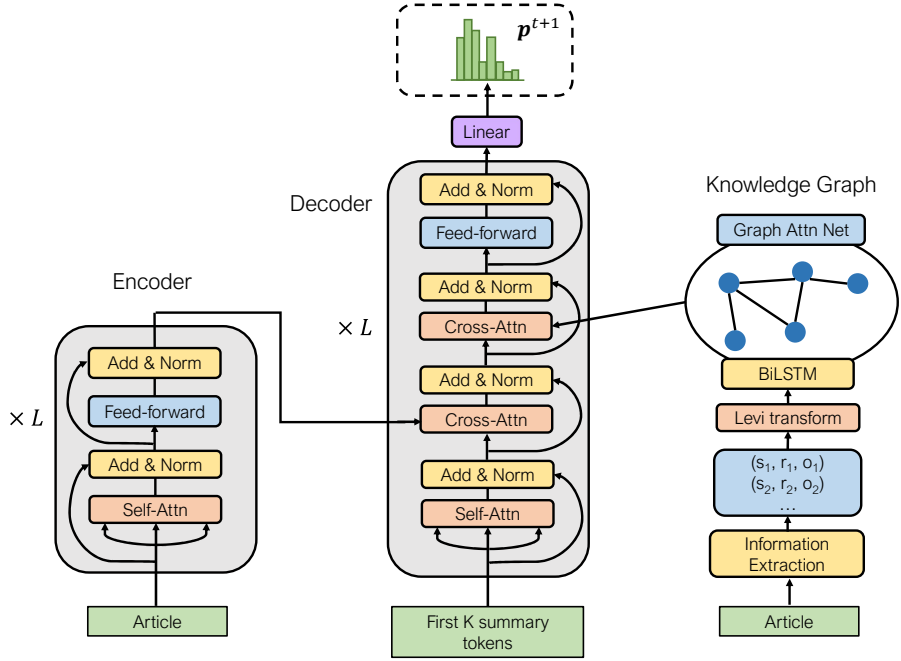}
\vspace{5.0\baselineskip}
\caption{The model architecture of \textsc{FASum}. It has $L$ layers of transformer blocks in both the encoder and decoder. The knowledge graph is obtained from information extraction results and it participates in the decoder's attention.}
\label{fig:fasum}
\end{figure*}

\subsection{Fact-Aware Summarizer}
We propose the Fact-Aware abstractive Summarizer, \textsc{FASum}. It utilizes the seq2seq architecture built upon transformers \citep{transformer}. In detail, the encoder produces contextualized embeddings of the article and the decoder attends to the encoder's output to generate the summary.

To make the summarization model fact-aware, we extract, represent and integrate knowledge from the source article into the summary generation process, which is described in the following. The overall architecture of \textsc{FASum} is shown in Figure~\ref{fig:fasum}.

\subsubsection{Knowledge Extraction}
To extract important entity-relation information from the article, we employ the Stanford OpenIE tool \citep{openie}. The extracted knowledge is a list of tuples. Each tuple contains a subject (S), a relation (R) and an object (O), each as a segment of text from the article. 
In the experiments, there are on average 165.4 tuples extracted per article in CNN/DailyMail \cite{cnn} and 84.5 tuples in XSum \cite{tconvs2s}.

\subsubsection{Knowledge Representation}
We construct a knowledge graph to represent the information extracted from OpenIE. We apply the Levi transformation \citep{levi} to treat each entity and relation equally. In detail, suppose a tuple is $(s, r, o)$, we create nodes $s$, $r$ and $o$, and add edges $s$---$r$ and $r$---$o$. In this way, we obtain an undirected knowledge graph $G=(V, E)$, where each node $v\in V$ is associated with text $t(v)$. During training, this graph G is constructed for each batch individually, i.e. there's no shared huge graph. One benefit is that the model can take unseen entities and relations during inference.

We then employ a graph attention network \citep{gat} to obtain embedding $\be_j$ for each node $v_j$. The initial embedding of $v_j$ is given by the last hidden state of a bidirectional LSTM applied to $t(v_j)$. In the experiment, we employ 2 graph attention layers.




\subsubsection{Knowledge Integration}
The knowledge graph embedding is obtained in parallel with the encoder. 
Then, apart from the canonical cross attention over the encoder's outputs, each decoder block also computes cross-attention over the knowlege graph nodes' embeddings:
\begin{align}
    &\alpha_{ij} = \mbox{softmax}_j(\beta_{ij}) = \frac{\mbox{exp}(\beta_{ij})}{\sum_{j\in V}\mbox{exp}(\beta_{ij})}\\
    &\beta_{ij} = \bs_i^T\be_j,\\
    &\bu_i = \sum_{j\in V}\alpha_{ij}\be_j,
\end{align}
where $\{\be_j\}_{j=1}^{|V|}$ are the final embeddings of the graph nodes, and $\{\bs_i\}_{i=1}^t$ are the decoder block's representation of the first $t$ generated tokens.


\subsubsection{Summary Generation}
We denote the final output of the decoder as  $\bz_1, ..., \bz_t$. To produce the next token $y_{t+1}$, we employ a linear layer $\bW$ to project $\bz_t$ to a vector of the same size of the dictionary. And the predicted distribution of $y_{t+1}$ is obtained by:
\begin{equation}
    \bp^{t+1}=\sigma(\bW \bz_t)
\end{equation}

During training, we use cross entropy as the loss function $\mathcal{L}(\theta) = -\sum_{t=1}^n \by_t^T \log(\bp^t)$, where $\by_t$ is the one-hot vector for the $t$-th token, and $\theta$ represent the parameters in the network.

\subsection{Fact Corrector}
To better utilize existing summarization systems, we propose a Factual Corrector model, \textsc{FC}, to improve the factual consistency of any summary generated by abstractive systems. \textsc{FC} frames the correction process as a seq2seq problem: given an article and a candidate summary, the model generates a corrected summary with minimal changes to be more factually consistent with the article.


While FASum has a graph attention module in the transformer, preventing direct adaptation from pre-trained models, the FC model architecture adopts the design of the pre-trained model UniLM \citep{unilm}. We initiailized the model weights from RoBERTa-Large \citep{roberta}. The finetuning process is similar to training a denoising autoencoder. We use back-translation and entity swap for synthetic data generation. For example, an entity in the ground-truth summary is randomly replaced with another entity of the same type from the article. This modified summary and the article is sent to the corrector to recover the original summary. In the experiments, we generated 3.0M seq2seq data samples in CNN/DailyMail and 551.0K samples in XSum for finetuning. We take 10K samples in each dataset for validation and use the rest for training.

During inference, the candidate summary from any abstractive summarization system is concatenated with the article and sent to FC, which produces the corrected summary.

\section{Experiments}
\label{sec:exp}
\subsection{Datasets}
We evaluate our model on benchmark summarization datasets CNN/DailyMail \cite{cnn} and XSum \cite{tconvs2s}. They contain 312K and 227K news articles and human-edited summaries respectively, covering different topics and various summarization styles.

\subsection{Implementation Details}
We use the Huggingface's \citep{huggingface} implementation of transformer in BART \citep{bart}. We also inherit their provided hyperparameters of CNN/DailyMail and XSum for the beam search. The minimum summary length is 56 and 11 for CNN/Daily Mail and XSum, respectively. The number of beams is 4 for CNN/DailyMail and 6 for XSum.

In \textsc{FASum}, both the encoder and decoder has 10 layers of 10 heads for attention. Teacher forcing is used in training. We use Adam \citep{adam} as the optimizer with a learning rate of 2e-4. 

The bi-LSTM to produce the initial embedding of graph nodes has a hidden state of size 64 and the graph attention network (GAT) has 8 heads and a hidden state of size 50. The dropout rate is 0.6 in GAT and 0.1 elsewhere. 

We use the subword tokenizer SentencePiece \cite{sentpiece}. The dictionary is shared across all the datasets. The vocabulary has a size of 32K and a dimension of 720.

The correction model \textsc{FC} follows the UniLM \citep{unilm} architecture initialized with weights from RoBERTa-Large \citep{roberta}. We fine-tune the model for 5 epochs with a learning rate of 1e-5 and linear warmup over the one-fifths of total steps and linear decay. During decoding, it uses beam search with a width of 2, and blocks tri-gram duplicates. The batch size during finetuning is 24. More details are presented in the Appendix.

\subsection{Metrics}
To evaluate factual consistency, we re-implemented and trained the FactCC model \citep{eval}. The model outputs a score between 0 and 1, where a higher score indicates better consistency between the input article and summary.
The training of FactCC is independent of our summarizer so no parameters are shared. More details are in the Appendix.

We also employ the standard ROUGE-1, ROUGE-2 and ROUGE-L metrics \cite{rouge} to measure summary qualities. These three metrics evaluate the accuracy on unigrams, bigrams and the longest common subsequence. We report the F1 ROUGE scores in all experiments. And the ROUGE-L score on validation set is used to pick the best model for both \textsc{FASum} and FC.



\subsection{Baselines}
The following abstractive summarization models are selected as baseline systems.
\textsc{TConvS2S} \cite{tconvs2s} is based on topic modeling and convolutional neural networks. 
\textsc{BottomUp} \cite{bottomup} uses a bottom-up approach to generate summarization. \textsc{UniLM} \cite{unilm} utilizes large-scale pretraining to produce state-of-the-art abstractive summaries. We train the baseline models when the predictions are not available in their open-source repositories.

\subsection{Results}
As shown in Table~\ref{tab:main}, our model \textsc{FASum}\footnote{We have put code and all the generated summaries of all models in the supplementary materials.} outperforms all baseline systems in factual consistency scores in CNN/DailyMail and is only behind \textsc{UniLM} in XSum. In CNN/DailyMail, \textsc{FASum} is 0.6\% higher than \textsc{UniLM} and 3.9\% higher than \textsc{BottomUp} in factual score. Statistical test shows that the lead is statistically significant with p-value smaller than 0.05. The higher factual score of \textsc{UniLM} among baselines corroborates the findings in \citet{factualfaithful} that pre-trained models exhibit better factuality. But our proposed knowledge graph component can help the train-from-scratch \textsc{FASum} model to excel in factual consistency. 

We conduct ablation study to remove the knowledge graph component from \textsc{FASum}, resulting in the \textsc{Seq2Seq} model. As shown, there is a clear drop in factual score: 2.8\% in CNN/DailyMail and 0.9\% in XSum. This proves that the constructed knowledge graph can help increase the factual correctess of the generated summaries.

It's worth noticing that the ROUGE metric does not always reflect the factual consistency, sometimes even showing an inverse relationship, a phenomenon observed in multiple studies \citep{kryscinski-etal-2019-neural,factualfaithful}. For instance, although \textsc{BottomUp} has 0.69 higher ROUGE-1 points than \textsc{FASum} in CNN/DailyMail, there are many factual errors in its summaries, as shown in the human evaluation. On the other hand, to make sure the improved factual correctness of our models is not achieved by simply copying insignificant information from the article, we conduct analysis on abstractiveness in Section~\ref{sec:novel} and human evaluation in Section~\ref{sec:human}.


Furthermore, the correction model \textsc{FC} can effectively enhance the factual consistency of summaries generated by various baseline models, especially when the original summary has a relatively low factual consistency. For instance, on CNN/DM, the factual score of \textsc{BottomUp} increases by 1.4\% after correction. On XSum, after correction, the factual scores increase by 0.2\% to 3.1\% for all baseline models. Interestingly, \textsc{FC} can also boost the factual consistency of our \textsc{FASum} model. Furthermore, the correction has a rather small impact on the ROUGE score, and it can improve the ROUGE scores of most models in XSum dataset.


We check and find that \textsc{FC} only makes modest modifications necessary to the original summaries. For instance, \textsc{FC} modifies 48.3\% of summaries generated by \textsc{BottomUp} in CNN/DailyMail. These modified summaries contain very few changed tokens: 94.4\% of these corrected summaries contain 3 or fewer new tokens, while the summaries have on average 48.3 tokens.

In the appendix of supplementary materials, we show several examples of summaries given by \textsc{FASum} and corrected by \textsc{FC} to demonstrate the improved factual consistency of summarization.

\begin{table*}[!h]
\centering
    \begin{tabular}{lllll} \thickhline
        Model & 100$\times$Fact Score & ROUGE-1 & ROUGE-2 & ROUGE-L\\
       \thickhline
       \rowcolor[gray]{0.95}\multicolumn{5}{l}{\textbf{CNN/DailyMail}}\\
       \thickhline
       \textsc{BottomUp}& 83.9 & 41.22 & 18.68 & 38.34 \\
       Corrected by \textsc{FC} & 85.3$^*$ ($\uparrow$1.4\%) & 40.95 & 18.37 & 37.86\\
       \hline
       \textsc{UniLM}& 87.2 & \textbf{43.33} & \textbf{20.21} & \textbf{40.51} \\
       Corrected by \textsc{FC} & 87.0 ($\downarrow$0.2\%) & 42.75 & 20.07 & 39.83\\
       \hline
       \textsc{Seq2Seq} & 85.0 & 41.03 & 18.04 & 37.93 \\
       \textsc{FASum}& 87.8$^*$ & 40.53 & 17.84 & 37.40 \\
       Corrected by \textsc{FC} & \textbf{88.1} ($\uparrow$0.3\%) & 40.38 & 17.67 & 37.23 \\
       \thickhline
       \rowcolor[gray]{0.95}\multicolumn{5}{l}{\textbf{XSum}}\\
       \thickhline
       \textsc{BottomUp}& 78.0 & 26.91 & 7.66 & 20.01 \\
       Corrected by \textsc{FC} & 78.9$^*$ ($\uparrow$0.9\%) & 28.21 & 8.00 & 20.69\\
       \hline
       \textsc{TConvS2S}& 79.8 & 31.89 & 11.54 & 25.75 \\
       Corrected by \textsc{FC} & 82.9$^*$ ($\uparrow$3.1\%) & 32.44 & 11.83 & 26.02\\
       \hline
       \textsc{UniLM} & 83.2 & 42.14 & \textbf{19.53} & 34.13 \\
       Corrected by \textsc{FC} & \textbf{83.4}   ($\uparrow$0.2\%)& \textbf{42.18} & \textbf{19.53} & \textbf{34.15}\\
       \hline
      \textsc{Seq2Seq} &  80.6 & 31.44 & 10.91 &  24.69\\
       \textsc{FASum}& 81.5 & 30.28 & 10.03 & 23.76\\
       Corrected by \textsc{FC} & 81.7 ($\uparrow$0.2\%) & 30.20 & 9.97 & 23.68\\
       \thickhline
    \end{tabular}
\caption{\label{tab:main} Factual consistency score and ROUGE scores on CNN/DailyMail and XSum test set. *p-value$<0.05$.} 
\end{table*} 

\begin{figure}[t]
\centering
\vspace{2.8\baselineskip}
\includegraphics[scale=0.27,trim=6cm 6cm 6cm 4cm]{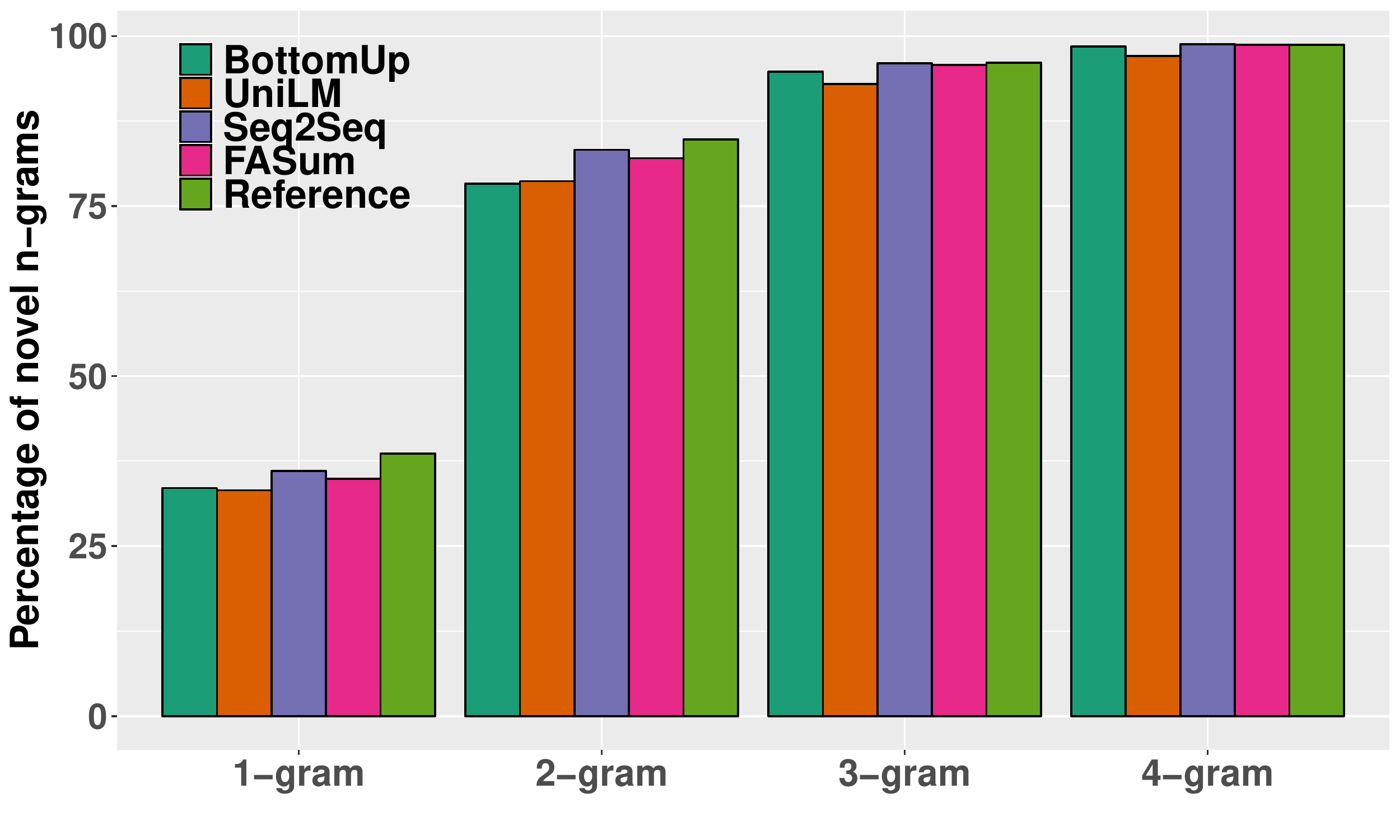} 
\vspace{3.2\baselineskip}
\caption{Percentage of novel n-grams for summaries in XSum test set.}
\label{fig:novel}
\end{figure}

\subsection{Insights}
\subsubsection{Novel n-grams}
\label{sec:novel}
It has been shown in \citet{feqa} that less abstractive summaries are more factual consistent with the article. Therefore, we inspect whether our models boost factual consistency simply by copying more portions of the article.

On XSum's testset, we compute the ratio of novel n-grams in summaries that do not appear in the article. Figure~\ref{fig:novel} shows that \textsc{FASum} achieves the closest ratio of novel n-gram compared with reference summaries, and higher than \textsc{BottomUp} and \textsc{UniLM}. 
This demonstrates that \textsc{FASum} can produce highly abstractive summaries while ensuring factual consistency.

\begin{table}[t]
\centering
    \begin{tabular}{l|c|c|c} \thickhline
        Model & $\mbox{RMR}_1\uparrow$ & $\mbox{RMR}_2\uparrow$ & $\mbox{NLI}\downarrow$ \\
        \hline
        \textsc{UniLM} & 60.0 & 39.6 & 10.2 \\
        \textsc{FC} correction & 61.4 & 40.7 & 10.0\\
        \hline
        \textsc{Seq2Seq} & 53.8 & 32.2 & 11.2 \\
       \textsc{FASum} & 65.0 & 46.0 & 9.3 \\
       \textsc{FC} correction & \textbf{67.0} & \textbf{47.4} & \textbf{8.3}\\
       \thickhline
    \end{tabular}
\caption{\label{tab:prec} Average relation matching rate (RMR, Eq.~\ref{eq:prec}) and NLI contradictory ratio between article and summary in CNN/DailyMail test set. The arrow indicates whether larger or smaller value means better result.} 
\end{table} 

\subsubsection{Relation Matching Rate}
While the factual consistency evaluator FactCC \citep{eval} is based on pre-trained models, it requires finetuning on articles and labelled summaries. Furthermore, we empirically find that the performance of FactCC degrades when it is finetuned on one summary dataset and used to evaluate models on another dataset.

Therefore, in this subsection, we design an easy-to-compute model-free factual consistency metric, which can be used when ground-truth summaries are not available.

As the relational tuples in the knowledge graph capture the factual information in the text, we compute the precision of extracted tuples in the summary. In detail, suppose the set of the relational tuples in the summary is $R_s=\{(s_i, r_i, o_i)\}$, and the set of the relational tuples in the article is $R_a$. Then, each tuple in $R_s$ falls into one of the following three categories:
\begin{enumerate}
    \item \textbf{Correct hit} (C): $(s_i, r_i, o_i)\in R_a$;
    \item \textbf{Wrong hit} (W): $(s_i, r_i, o_i)\not\in R_a$, but $\exists o'\neq o_i, (s_i, r_i, o')\in R_a$, or $\exists s'\neq s_i, (s', r_i, o_i)\in R_a$;
    \item \textbf{Miss} (M): Otherwise.
\end{enumerate}

We define two kinds of relation matching rate (RMR) to measure the ratio of correct hits: 
\begin{align}
    \mbox{RMR}_1&=100\times\frac{C}{C+W}\\
    \mbox{RMR}_2&=100\times\frac{C}{C+W+M}
\label{eq:prec}
\end{align}

Note that this metric is different from the ratio of overlapping tuples proposed in \citet{assessfactual}, where the ratio is computed between the ground-truth and the candidate summary. Since even the ground-truth summary may not cover all the salient information in the article, we choose to compare the knowledge tuples in the candidate summary directly against those in the article. An additional advantage of our metric is that it does not require ground-truth summaries to be available.

Table~\ref{tab:prec} displays the result of this metric in CNN/DailyMail's testset. As shown, \textsc{FASum} achieves the highest precision of correct hits under both measures. And there is a considerable boost from the knowledge graph (\textsc{FASum} vs \textsc{Seq2Seq}): 11.2\% in $\mbox{RMR}_1$ and 13.8\% in $\mbox{RMR}_2$. And the correction from the FC model can further improve the metric for both \textsc{FASum} and \textsc{UniLM}. 

We also compute factual consistency via natural language inference models following \citet{factualfaithful}. We use the BERT-Large model finetuned on MNLI dataset \citep{mnli} provided by fairseq \citep{fairseq}. The model predicts the relationship between the article and summary to be one of the following: entailment, neutral and contradiction. We report the ratio of contradiction as predicted by the model in Table~\ref{tab:prec}. As shown, \textsc{FASum} achieves the lowest ratio and \textsc{FC} helps further reducing conflicting facts in generated summaries.

\subsubsection{Human Evaluation} 
\label{sec:human}
\begin{table}[t]
\centering
    \begin{tabular}{l|c|c} \thickhline
        Model & Factual Score & Informativeness\\
        \hline
        \textsc{BottomUp} & 2.32 & 2.23 \\
        \hline
        \textsc{UniLM} & 2.65 & \textbf{2.45} \\
        \hline
        \textsc{Seq2Seq} & 2.59 & 2.30 \\
       \textsc{FASum} & \textbf{2.74}$^*$ & 2.42 \\
       \thickhline
    \end{tabular}
\caption{\label{tab:human} Human evaluation results of summaries for 100 randomly sampled articles in CNN/DailyMail test set. *p-value$<0.05$.} 
\end{table} 

\begin{table}[!t]
\centering
    \begin{tabular}{c|c|c} \thickhline
        \textsc{BottomUp} is better & \textsc{FC} is better & Same\\
        \hline
        15.0\% & 42.3\% & 42.7\% \\
        \hline
        \textsc{UniLM} is better & \textsc{FC} is better & Same\\
        \hline
        20.4\% & 31.2\% & 48.4\%\\
       \thickhline
    \end{tabular}
\caption{\label{tab:human-side-by-side} Human evaluation results for side-by-side comparison of factual consistency on 100 randomly sampled articles from CNN/DailyMail test set where \textsc{FC} makes modifications.} 
\end{table} 

We conduct human evaluation on the factual consistency and informativeness of summaries. We randomly sample 100 articles from the test set of CNN/DailyMail. Then, each article and summary pair is labelled by 3 people from Amazon Mechanical Turk (AMT) to evaluate the factual consistency and informativeness. Each labeller gives a score in each category between 1 and 3 (3 being perfect). The kappa-ratio between reviewer scores is 0.32 for factual consistency and 0.28 for informativeness.

Here, factual consistency indicates whether the summary's content is faithful with respect to the article; informativeness indicates how well the summary covers the salient information in the article.

As shown in Table~\ref{tab:human}, our model \textsc{FASum} achieves the highest factual consistency score, higher than \textsc{UniLM} and considerably outperforming \textsc{BottomUp}. We conduct a statistical test and find that compared with \textsc{UniLM}, our model's score is statistically significant with p-value smaller than 0.05 under paired t-test. In terms of informativeness, our model is comparable with \textsc{UniLM} and outperforms \textsc{BottomUp}. Finally, without the knowledge graph component, the \textsc{Seq2Seq} model generates summaries with both less factual consistency and informativeness.

To assess the effectiveness of the correction model \textsc{FC}, we conduct a human evaluation of side-by-side summaries. In CNN/DailyMail, we randomly sample 100 articles where the summaries generated by \textsc{BottomUp} are modified by \textsc{FC}. 3 labelers are asked whether the original or the corrected version is factually more correct. We collect all the feedbacks and compute the ratio of judgements for each case. To reduce bias, we randomly shuffle the two versions of summaries. We conduct similar evaluation on \textsc{UniLM}.

As shown in Table~\ref{tab:human-side-by-side}, the corrected summaries are significantly more likely to be judged as more factually correct for both baseline models. For example, 42.3\% of the judgements think the corrected summaries are factually more correct, 42.7\% think the corrected version neither improves nor worsens the factual consistency, while only 15.0\% think that the corrected version becomes worse than the original \textsc{BottomUp} summary. Therefore, FC can help boost the factual consistency of summaries from given systems.

Finally, to evaluate the quality of the relation matching rate (RMR), we compute the correlation coefficient $\gamma$ between the factual score given by human labelers and the RMR value. The result shows that $\gamma=0.43$, indicating observable relationship between RMR and human evaluation results.


\section{Conclusion}
\label{sec:conclusion}
In this paper, we extract factual information from the article to be represented by a knowledge graph. We then integrate this factual knowledge into the process of producing summaries. The resulting model \textsc{FASum} enhances the ability to preserve facts during summarization, demonstrated by both automatic and human evaluation. We also present a correction model, \textsc{FC}, to rectify factual errors in candidate summaries. Furthermore, we propose an easy-to-compute model-free metric, relation matching rate, to measure factual consistency based on the overlapping ratio of relational tuples.

For future work, we plan to integrate knowledge graphs into pre-training for more accurate and factually consistent  summarization. Moreover, we will combine the internally extracted knowledge graph with an external knowledge graph (e.g. ConceptNet) to enhance the commonsense capability of summarization models.

\bibliography{main}
\bibliographystyle{acl_natbib}

\newpage
\clearpage
\appendix
\section{Implementation details}
For hyperparameter search, we tried 4 layers with 4 heads, 6 layers with 6 heads and 10 layers with 10 heads. 

There're 108.3M parameters in the \textsc{FASum} model and it takes 2 hours (CNN/DailyMail) / 0.5 hours (XSum) for 4 v100 GPUs to train 1 epoch. The batch size is set to 48 for both datasets.

On validation datasets, \textsc{FASum} achieves ROUGE-1 41.08\%, ROUGE-2 18.35\% and ROUGE-L 37.95\% on CNN/DailyMail, and it achieves ROUGE-1 30.28\%, ROUGE-2 10.09\% and ROUGE-L 23.85\% on XSum.


\section{Factual Consistency Evaluator}
\label{sec:eval}
To automatically evaluate the factual consistency of a summary, we leverage the FactCC model \citep{eval}, which maps the consistency evaluation as a binary classification problem, namely finding a function $f:(A, C)\longrightarrow[0,1]$, where $A$ is an article and $C$ is a summary sentence defined as a claim. $f(A, C)$ represents the probability that $C$ is factually correct with respect to the article $A$. If a summary $S$ is composed of multiple sentences $C_1, ..., C_k$, we define the factual score of $S$ as: $f(A, S)=\frac{1}{k}\sum_{i=1}^k f(A, C_i)$.
 
\begin{table}[t]
\centering
    \begin{tabular}{l|c} \thickhline
        Model & Incorrect \\
        \hline
        Random & 50.0\% \\
        \hline
BERT \citep{ranking} & 35.9\% \\
ESIM \citep{ranking} & 32.4\% \\
\hline
FactCC \citep{eval} & 30.0\% \\
FactCC (our version) & \textbf{26.8\%}\\
       \thickhline
    \end{tabular}
\caption{\label{tab:evaluator} Percentage of incorrectly ordered sentence pairs using different consistency prediction models in CNN/DailyMail, using data from \citet{ranking}.} 
\end{table} 
 
To generate training data, we adopt backtranslation as a paraphrasing tool. The ground-truth summary is translated into an intermediate language, including French, German, Chinese, Spanish and Russian, and then translated back to English. Together with the original summaries, these claims are used as positive training examples. We then apply entity swap, negation and pronoun swap to generate negative examples \citep{eval}.
 
 
 
Following \citet{eval}, we finetune the BERT\textsubscript{BASE} model using the same hyperparameters to finetune FactCC. We concatenate the article and the generated claim together with special tokens [CLS] and [SEP]. The final embedding of [CLS] is used to compute the probability that the claim is entailed by the article content.
 
As shown in Table~\ref{tab:evaluator}, on CNN/Daily Mail, our reproduced model achieves better accuracy than that in \citet{eval} on the human-labelled sentence-pair-ordering data \citep{ranking}. Thus, we use this evaluator for all the factual consistency assessment tasks in the following.\footnote{We use the same setting and train another evaluator for XSum dataset.}

\section{Examples}
Table~\ref{table:example1}, \ref{table:example2} and \ref{table:example3} show examples of CNN/DailyMail articles and summaries generated by our model and several baseline systems. The factual errors in the summary are marked in red, the correct facts in the summaries of \textsc{FASum} are marked in green and the corresponding facts are marked in bold in the article.

As shown, while baseline systems like \textsc{BottomUp} and \textsc{UniLM} achieve high ROUGE scores, they are susceptible to factual errors. For instance, in Article 5, both \textsc{BottomUp} and \textsc{Seq2Seq} wrongly state that Rickie Fowler accused Alexis. In fact, Alexis, Rickie's girlfriend, was accused by an online hater. In Article 1, \textsc{UniLM} mistakenly summarizes that Arsenal lost 4-1 where in fact Arsenal won 4-1 against Liverpool.

In comparison, our proposed fact-aware summarizer \textsc{FASum} could faithfully summarize the salient information in the article. And it can re-organize the phrasing instead of merely copying content from the article. 

Table~\ref{table:fc_example1} and Table~\ref{table:fc_example2} show examples of CNN/DailyMail articles, summaries generated by \textsc{BottomUp}/\textsc{UniLM} and the corrected version by \textsc{FC}. As shown, our correction model can select the wrong entities and replace them with correct ones. For instance, in Article 1, \textsc{BottomUp}'s summary states that Rual Castro, who appears elsewhere in the article, is the President of Venezuela, while \textsc{FC} correctly replaces it with Nocolas Maduro. In Article 4, \textsc{UniLM} wrongly attributes the statement to Scott's lawyer (probabily because ``Scott'' appears closer to the statement in the article), while it was actually said by Slager's lawyer. This error is corrected by \textsc{FC}.

\begin{table*}[htbp]
\centering
\begin{tabular}{m{2cm}|m{12.7cm}}
\thickerhline
Article 1 & Arsenal target Raheem Sterling is \textbf{a more talented player} than Gunners star Mesut Ozil, according to Michael Owen. \textbf{The former Liverpool forward made the claim} before Brendan Rodgers' men lost 4-1 at the Emirates on Saturday ... Ozil bends home Arsenal's second goal in their \textbf{4-1 win against Liverpool} at the Emirates... \\
\hline
\hline
Reference & Arsenal are interested in signing Liverpool's Raheem Sterling this summer. The 20-year-old has rejected a new PS100,000-a-week deal at Anfield. Michael Owen believes \textbf{Sterling is more talented than Mesut Ozil}. Adrian Durham: Sterling would be earning the same as Balotelli if he signed PS100,000-a-week deal at Liverpool... Durham: Arsenal only turn it on when the pressure is off. The German scored in \textbf{Arsenal's 4-1 win} against Liverpool on Saturday... \\
\hline
\textsc{BottomUp} & Raheem Sterling made the claim before Brendan Rodgers' men lost 4-1 at the Emirates on Saturday.	Bt Sport pundit Owen believes there should not be a debate about whether Sterling is more talented than Ozil.	\textbf{\textcolor{fireenginered}{The former Liverpool forward}} is a better player than Mesut Ozil.\\
\hline
\textsc{UniLM} &  Michael Owen believes Raheem Sterling is a better player than Mesut Ozil. Owen made the claim before \textbf{\textcolor{fireenginered}{Arsenal lost 4-1}} at the Emirates on Saturday. Owen's opinion drew criticism on social media but the former Liverpool forward has refused to back down. Owen said :` Is there a debate? Not even close' CLICK HERE for all the latest Arsenal news. \\
\hline
\textsc{FASum} &  Michael Owen says there \textbf{\textcolor{forestgreen}{should not be a debate}} about whether Raheem Sterling is more talented than Mesut Ozil.  The former Liverpool forward made the claim before \textbf{\textcolor{forestgreen}{Brendan Rodgers' men lost 4-1}} at the Emirates on Saturday.  Owen repeated his claim that the wantaway forward is a superior player than Ozil made.\\
\hline
\textsc{Seq2Seq} & Michael Owen made the claim before Liverpool's men lost 4-1 at the Emirates on Saturday. \textbf{\textcolor{fireenginered}{Michael Owen believes there should be a debate}} about whether Sterling is more talented than Ozil. The ex-England striker has refused to back down, even after Sterling's poor display in north London. \\
\thickerhline
Article 2 & \textbf{Yaya Toure was accused by Jamie Carragher} of ducking out of the way of Jason Puncheon's free-kick which killed off Manchester City's hopes of salvaging something from Selhurst Park. Toure, who is \textbf{paid PS200,000-a-week}, was \textbf{part of a five-man wall} when \textbf{Puncheon stepped up} to double Crystal Palace's lead...  \\
\hline
\hline
Reference & Crystal Palace beat Manchester City 2-1 at Selhurst Park on Monday night. Jason Puncheon's free-kick \textbf{doubled Palace's lead} in the second half. \textbf{Yaya Toure} was accused of ducking out of the way of the winning goal. Sportsmail's \textbf{Jamie Carragher:} 'The rest of them are desperate to be hit with the ball in the face and it's poor from Yaya Toure' Gary Neville: City have a mentality problem... they cannot sustain success. \\
\hline
\textsc{BottomUp} & Yaya Toure was accused by Jamie Carragher out of the way of Jason Puncheon's free-kick at Selhurst Park on Monday.	Toure, who is paid 200,000-a-week, was part of a \textbf{\textcolor{fireenginered}{wall with Crystal Palace's lead.}}	Carragher said the midfielder did not do enough to prevent the winner.\\
\hline
\textsc{UniLM} & Yaya Toure was accused by Sportsmail's Jamie Carragher of ducking out of the way of Jason Puncheon's free-kick at Selhurst Park on Monday. The midfielder was part of a five-man wall when Puncheon stepped up to double Crystal Palace's lead. Toure is paid 200,000-a-week a \textbf{\textcolor{fireenginered}{by the Premier League champions}}.\\
\hline
\textsc{FASum} & \textbf{\textcolor{forestgreen}{Yaya Toure was accused by Jamie Carragher}} of ducking out of the way of Jason Puncheon's free-kick.  Fernandinho and Vincent Kompany look keen to get their head to the ball but Toure appears to duck.  \textbf{\textcolor{forestgreen}{Toure paid PS200,000-a-week}}, was \textbf{\textcolor{forestgreen}{part of a five-man wall}} when Puncheon stepped up to \textbf{\textcolor{forestgreen}{double Crystal Palace's lead.}}\\
\hline
\textsc{Seq2Seq} & \textbf{\textcolor{fireenginered}{Jamie Carragher was accused}} of ducking out of the way of Jason Puncheon's free-kick which killed off Manchester City's chances at Selhurst Park on Monday Night. Toure is paid PS200,000-a-week, was part of a five-man wall when Puncheon stepped up to double Palace's lead. Sportsmail columnist Jamie Carragher feels City midfielder did not do enough to prevent winner. \\
\thickerhline
\end{tabular}
\caption{Sample articles and summaries from our model \textsc{FASum} and baseline systems in CNN/DailMail. \textsc{Seq2Seq} indicates our model without the knowledge graph component. The factual errors in summaries are marked in red, the correct facts in \textsc{FASum} are marked in green and the corresponding facts are marked in bold in the article.}
\label{table:example1}
\end{table*}


\begin{table*}[htbp]
\centering
\begin{tabular}{m{2.0cm}|m{12.7cm}}
\thickerhline
Article 3 & Real Madrid ace Gareth Bale treated himself to a Sunday evening BBQ after ... a sensational victory. The Welsh wizard was ... \textbf{scoring twice} and assisting another in an impressive victory in Israel... he \textbf{opened the scoring in Real Madrid's impressive 9-1 victory}... His goal in the \textbf{25th minute} may have got the ball rolling ... \textbf{Cristiano Ronaldo scored five goals} against Granada on Sunday ... 
\\
\hline
\hline
Reference &  Welsh Wizard Gareth Bale opened the scoring in Sunday's stunning win. Real Madrid put nine goals past Granada to keep pressure on Barcelona. Bale treated himself to a BBQ in the Spanish sun following victory. Cristiano Ronaldo scored five goals in the sensational team performance.\\
\hline
\textsc{BottomUp} & Gareth Bale scored twice in Real Madrid's 9-1 victory against Granada. \textbf{\textcolor{fireenginered}{The Real Madrid ace}} scored five goals against Granada on Sunday. The Welsh wizard was in impressive form for the international break.
\\
\hline
\textsc{UniLM} & Gareth Bale scored twice as Real Madrid beat Granada 9-1. The Welsh winger scored for Wales away on international duty.  Bale Tweeted:`Unbelievable result this afternoon, great team performance. Now for a BBQ' Real Madrid bounced back from El Clasico defeat by Barcelona with a thumping win. Cristiano Ronaldo scored five goals against an eight-minute hat-trick.\\
\hline
\textsc{FASum} &   Real Madrid beat Granada 9-1 at the Bernabeu on Sunday evening.  \textbf{\textcolor{forestgreen}{Gareth Bale scored twice}} and helped his side to a sensational 9-1 win.  \textbf{\textcolor{forestgreen}{Cristiano Ronaldo scored five goals}} against Granada on Sunday.  The Welsh winger Tweeted:'Unbelievable result this afternoon, great team performance now for a BBQ'\\
\hline
\textsc{Seq2Seq} & Gareth Bale scored twice as Real Madrid beat Granada 9-1 at the Bernabeu on Sunday. The Welsh ace opened the scoring with a goal against Granada in the 25th minute. \textbf{\textcolor{fireenginered}{Gareth Bale scored five}} and assisted another in an impressive win in Israel. Bale Tweeted:' Unbelievable result this afternoon, great team performance. Now for a BBQ' \\

\thickerhline
Article 4 & \textbf{\textit{Former} Huddersfield centre} \textbf{Josh Griffin scored a try and kicked three goals} as improving \textbf{Salford secured back-to-back Super League wins} with an 18-12 triumph at the John Smith's Stadium. \textbf{Griffin and Ben Jones-Bishop scored tries} in the first half, although Jack Hughes gave the Giants hope. \textbf{Carl Forster's try and Griffin's penalty} made the game safe... \textbf{Salford Red Devils} secured a second win?...\\
\hline
\hline
Reference & Salford sealed consecutive Super League wins with victory at Huddersfield. Josh Griffin inspired Red Devils with a try and three kicked goals. Ben Jones-Bishop and Carl Forster scored the other tries for the visitors.\\
\hline
\textsc{BottomUp} & Josh Griffin and Ben Jones-Bishop scored tries in the first half.	Salford Red Devils win a second win in a row with a 18-12 victory at Huddersfield.	\textbf{\textcolor{fireenginered}{Carl Forster scored a try and kicked three goals}} for Salford. \\
\hline
\textsc{UniLM} & Former Huddersfield centre Josh Griffin scored a try and kicked three goals. Ben Jones-Bishop scored tries in the first half. Carl Forster's try and Griffin's penalty made the game safe.\\
\hline
\textsc{FASum} & \textbf{\textcolor{forestgreen}{Former Huddersfield centre Josh Griffin scored a try and kicked three goals}} as Salford secured back-to-back Super League wins with an 18-12 win at John Smith's Stadium.  Ben Jones-Bishop scored tries in the first half but Jack Hughes gave the Giants hope.  \textbf{\textcolor{forestgreen}{Carl Forster's try and Griffin's penalty}} made the game safe, though.\\
\hline
\textsc{Seq2Seq} & \textbf{\textcolor{fireenginered}{Huddersfield centre}} Josh Griffin scored a try and kicked three goals. Griffin scored tries in the first half for the Giants. The game was twice delayed early on when \textbf{\textcolor{fireenginered}{Kevin Locke for Huddersfield's}} Brett Ferres were carried off injured.\\
\thickerhline
\end{tabular}
\caption{Continuation of Table~\ref{table:example1}.}
\label{table:example2}
\end{table*} 

\begin{table*}[htbp]
\centering
\begin{tabular}{m{2.0cm}|m{12.7cm}}
\thickerhline
Article 5 & \textbf{Rickie Fowler has earned extra brownie points} with his growing legion of fans after calling out an online hater who accused his girlfriend of being a `gold digger.' The golfer's girlfriend, bikini model \textbf{Alexis Randock, had posted a photo} on her Instagram account of her and sister Nicole on the beach last week... an \textbf{abusive commenter known as 'fatalsplash'} who was quick to judge and accused Alexis of being a `gold digger.' ... Fowler told the hater to \textbf{'get your facts straight'} ...\\
\hline
\hline
Reference & Rickie Fowler responded to  online troll who abused his girlfriend. Bikini model Alexis Randock had posted a photo on her Instagram account. Randock was pictured on the beach alongside her sister Nicole. A troll called `fatalsplash' then accused Alexis of being a `gold digger' Fowler told the hater to 'get your facts straight' following comment.\\
\hline
\textsc{BottomUp} & \textbf{\textcolor{fireenginered}{Rickie Fowler accused Alexis}} of being a `gold digger' who did n't have to work.	 \textbf{\textcolor{fireenginered}{The golfer was criticised}} by an online troll for not working due to her relationship with Fowler.	Fowler has earned extra brownie points with his legion of fans. \\
\hline
\textsc{UniLM} & Rickie Fowler 's girlfriend, bikini model Alexis Randock, posted a photo on her Instagram account of her and sister Nicole on the beach last week. An abusive commenter called `fatalsplash' accused Alexis of being a `gold digger' \\
\hline
\textsc{FASum} & Rickie Fowler has earned extra brownie points with his growing legion of fans.  The golfer's girlfriend, bikini model \textbf{\textcolor{forestgreen}{Alexis Randock, had posted a photo}} on her Instagram account.  The photo was shared with her 25,000 followers, which included an abusive commenter known as `fatalsplash' who was quick to judge and accused Alexis of being a `gold digger' \\
\hline
\textsc{Seq2Seq} & Golfer's girlfriend posted a picture of her and sister Nicole Randock on Instagram. The picture shared with her 25,000 followers was\textbf{\textcolor{fireenginered}{shared by her sister}} on the beach last week. The  \textbf{\textcolor{fireenginered}{golfer was quick to judge and accused Alexis}} of being a  \textbf{\textcolor{fireenginered}{'gold digger' troll}}. He responded by \textbf{\textcolor{fireenginered}{posting the caption}} `fatalsplash' accusing her of being 'a digger who didn't work for it' \\

\thickerhline
Article 6 & The VII Summit of the Americas was supposed to be all about the symbolic handshake between the United States and Cuba... 
Heads of state from 35 countries in the Western Hemisphere have met every three years...
\textbf{Obama and Castro would come face-to-face}... \textbf{Venezuelan President Nicolas Maduro} recently accused the United States of trying to topple his government...\\
\hline
\hline
Reference & U.S., Venezuelan relations threaten to overshadow Obama, Castro meeting. Venezuelan President says United States moved to oust him; he has the support of the Cuban foreign minister.\\
\hline
\textsc{BottomUp} & Heads of state from 35 countries have met every three years to discuss economic, social or political issues since the summit in 1994.	\textbf{\textcolor{fireenginered}{Venezuela's President Raul Castro}} has been criticized for human rights violations.	The \textbf{\textcolor{fireenginered}{U.S. government says}} the summit of the Americas is a ``triangle of tension.'' \\
\hline
\textsc{UniLM} & Summit of the Americas was supposed to be all about the symbolic handshake between the U.S. And Cuba. \textbf{\textcolor{fireenginered}{The tide changed when}} President Obama and Castro announced that more than five decades of Cold War rivalry was ending. Venezuelan President Nicolas Maduro recently accused the U.S. Of trying to topple his government...\\
\hline
\textsc{FASum} & The summit of the Americas was supposed to be all about the symbolic handshake between the u.s. and Cuba.  Diplomats from both countries began negotiations \textbf{\textcolor{forestgreen}{to establish embassies in Havana and Washington}}.  President Nicolas Maduro recently accused the United States of trying to topple his government.\\
\hline
\textsc{Seq2Seq} & The VII summit of the Americas was supposed to be all about the symbolic handshake between the United States and Cuba. Relations between the u.s. and Cuba have been the wrench in diplomatic machinery, with some Latin American leaders threatening not to attend. President Barack Obama and Castro \textbf{\textcolor{fireenginered}{face-to-face embassies}} in Havana and Washington.\\

\thickerhline

\end{tabular}
\caption{Continuation of Table~\ref{table:example1}.}
\label{table:example3}
\end{table*}

\begin{table*}[htbp]
\centering
\begin{tabular}{m{2.0cm}|m{12.7cm}}
\thickerhline
Article 1 & The VII Summit of the Americas was supposed to be all about the symbolic handshake between the United States and Cuba. But insert Venezuela into the mix and Panama City, Panama, quickly turns into a ``triangle of tension.''... 
Cuba has historically been the wrench in the diplomatic machinery, with some Latin American leaders threatening not to attend the Summit of the Americas if the United States and Canada didn't agree to invite President Raul Castro...
The much anticipated handshake between Obama and Castro would steal all the headlines if it wasn't for Cuba's strongest ally, Venezuela. Venezuelan President \textbf{Nicolas Maduro} recently accused the United States of trying to topple his government and banned former President George Bush... \\
\hline
\hline
\textsc{BottomUp} & Heads of state from 35 countries have met every three years to discuss economic, social or political issues since the summit in 1994. Venezuela's President \textbf{\textcolor{fireenginered}{Raul Castro}} has been criticized for human rights violations. The u.s. government says the summit of the Americas is a ``triangle of tension.'' \\
\hline
Corrected by \textsc{FC} & Heads of state from 35 countries have met every three years to discuss economic, social or political issues since the summit in 1994. Venezuela's President \textbf{\textcolor{forestgreen}{Nicolas Maduro}} has been criticized for human rights violations. The u.s. government says the summit of the Americas is a ``triangle of tension.''\\
\hline
\thickerhline
Article 2 & She's one of the hottest and most successful Latinas in Hollywood, but now Sofia Vergara is playing defense in a legal battle initiated by her ex-fiance: He wants to keep the two frozen embryos from their relationship, both female. The 42-year-old actress and star of the hit TV sitcom "Modern Family" split from businessman \textbf{Nick Loeb} in May 2014. Loeb is suing the Colombian-born actress in Los Angeles to prevent Vergara from destroying their two embryos conceived through in vitro fertilization in November 2013, according to published reports by New York Daily News and In Touch magazine...\\
\hline
\hline
\textsc{BottomUp} & \textbf{\textcolor{fireenginered}{Sofia Vergara}} wants to keep the frozen embryos from their relationship, both female. He is suing the actress in Los Angeles to prevent Vergara from their embryos. The actress and star of the ``Modern Family'' split from Nick Loeb in May 2014.\\
\hline
Corrected by \textsc{FC} & \textbf{\textcolor{forestgreen}{Nick Loeb}} wants to keep the frozen embryos from their relationship, both female. He is suing the actress in Los Angeles to prevent Vergara from their embryos. The actress and star of the``Modern Family"split from Businessman Nick Loeb in May 2014.\\
\thickerhline
Article 3 & Volvo says it will begin exporting vehicles made in a factory in southwest China to the United States next month, the first time Chinese-built passenger cars will roll into American showrooms. Parent company \textbf{Geely Automobile}, which bought Volvo in 2010, is seeking to prove that a Chinese company can manage a global auto brand...\\
\hline
\hline
\textsc{BottomUp} & \textbf{\textcolor{fireenginered}{Volvo}} is seeking to prove that a Chinese company can manage a global auto brand. The car will be one of four models produced in a manufacturing plant in Chengdu. China is largest market for car sales globally in 2009.\\
\hline
Corrected by \textsc{FC} & 
\textbf{\textcolor{forestgreen}{Geely Automobile}} is seeking to prove that a Chinese company can manage a global auto brand. The car will be one of four models produced in a manufacturing plant in Chengdu. China is largest market for car sales globally in 2009.\\
\thickerhline

\end{tabular}
\caption{Example articles, summaries and corrections made by \textsc{FC} in CNN/DailMail. The factual errors in the original summary are marked in red, the corrected part by \textsc{FC} are marked in green and the corresponding facts are marked in bold in the article.}
\label{table:fc_example1}
\end{table*}


\begin{table*}[htbp]
\centering
\begin{tabular}{m{2.0cm}|m{12.7cm}}
\thickerhline
Article 4 & ... Scott family attorney Chris Stewart said the man with Scott was a co-worker and friend... Slager has been fired and faces life in prison or the death penalty if convicted on a murder charge. Video shot by a witness shows the officer shooting Scott in the back as he runs away. \textbf{Slager} had told investigators he feared for his safety after he and Scott tussled over his Taser. \textbf{His lawyer}, Andy Savage, said Friday he ``has not received the cooperation from law enforcement that the media has.'' ...\\
\hline
\hline
\textsc{UniLM} & ``This is a heartbreaking tragedy for everyone in our community,'' Charleston mayor says. Scott family attorney says the man with Scott was a co-worker and friend. Michael Slager has been fired and faces life in prison or the death penalty if convicted on a murder charge. \textbf{\textcolor{fireenginered}{Scott}}'s lawyer says he ``has not received the cooperation from law enforcement that the media has  \\
\hline
Corrected by \textsc{FC} & ``This is a heartbreaking tragedy for everyone in our community,'' Charleston mayor says. Scott family attorney says the man with Scott was a co-worker and friend. Michael Slager has been fired and faces life in prison or the death penalty if convicted on a murder charge. \textbf{\textcolor{forestgreen}{Slager}}'s lawyer says he ``has not received the cooperation from law enforcement that the media has\\
\thickerhline
Article 5 & \textbf{Rangers boss Stuart McCall} has revealed he was warned about taking over at Ibrox by Walter Smith - but insists he has made the right decision. The former Light Blues midfielder steered his new side to their third victory on the bounce with Sunday's 2-1 win over Hearts. It was a crucial triumph that moves \textbf{Gers} back into second place in the Scottish Championship and looks to have restored vital momentum ahead of the end-of-season play-offs. \textbf{Stuart McCall says} he was warned not to take the Rangers job but is pleased that he took the risk in the end.  \\ 
\hline
\hline
\textsc{UniLM} & Stuart McCall guided Rangers to a 2-1 win over Hearts on Sunday. Kenny Miller and Haris Vuckic scored the goals for the Gers at Ibrox. The win moved \textbf{\textcolor{fireenginered}{Rangers}} back into second place in the Scottish Championship. The \textbf{\textcolor{fireenginered}{Gers}} boss says he was warned not to take over by Walter Smith. But he says he is pleased that he took the risk in the end.\\
\hline
Corrected by \textsc{FC} & Stuart McCall guided Rangers to a 2-1 win over Hearts on Sunday. Kenny Miller and Haris Vuckic scored the goals for the Gers at Ibrox. The win moved \textbf{\textcolor{forestgreen}{Gers}} back into second place in the Scottish Championship. The \textbf{\textcolor{forestgreen}{Rangers}} boss says he was warned not to take over by Walter Smith. But he says he is pleased that he took the risk in the end.\\
\thickerhline
Article 6 & \textbf{Wasps} will appeal against the three-week ban issued to \textbf{Nathan Hughes}, but their protest will not enable the back row to play in Sunday's Champions Cup quarter-final against Toulon. The club have said in a statement they will appeal on the grounds the incident that left \textbf{Northampton wing George North} unconscious and resulted in a red card for Hughes was an accident and no foul play had occurred... \textbf{Wasps} are set to appeal the three-match ban handed to No 8 \textbf{Nathan Hughes} for knocking out \textbf{George North}... \textbf{Hughes} was given the suspension - to include three meaningful fixtures - after being \textbf{found guilty} of striking North in the 52-30 Aviva Premiership defeat at Franklin's Gardens on March 27...\\
\hline
\hline
\textsc{UniLM} & Wasps will appeal the three-match ban handed to No 8 Nathan Hughes. The \textbf{\textcolor{fireenginered}{Northampton}} winger was found guilty of striking George North. The incident occurred during \textbf{\textcolor{fireenginered}{Northampton}}'s 52-30 defeat at Franklin's Gardens. North suffered a third confirmed concussion in just four months.\\ 
\hline
Corrected by \textsc{FC} & Wasps will appeal the three-match ban handed to No 8 Nathan Hughes. The \textbf{\textcolor{forestgreen}{Wasps}} winger was found guilty of striking George North. The incident occurred during \textbf{\textcolor{forestgreen}{Wasps}}'s 52-30 defeat at Franklin's Gardens. North suffered a third confirmed concussion in just four months. \\

\thickerhline

\end{tabular}
\caption{Continuation of Table~\ref{table:fc_example1}}
\label{table:fc_example2}
\end{table*}
\end{document}